\documentclass[10pt,twocolumn,letterpaper]{article}

\usepackage{cvpr}              % To produce the CAMERA-READY version
\usepackage{amssymb}

\usepackage[utf8]{inputenc} % allow utf-8 input
\usepackage[T1]{fontenc}    % use 8-bit T1 fonts
\usepackage{url}            % simple URL typesetting
\usepackage{booktabs}       % professional-quality tables
\usepackage{amsfonts}       % blackboard math symbols
\usepackage{nicefrac}       % compact symbols for 1/2, etc.
\usepackage{microtype}      % microtypography
\usepackage[table]{xcolor}

\usepackage{graphicx}
\usepackage{multirow}
\usepackage{bbding}

\definecolor{hollywoodcerise}{rgb}{0.96, 0.0, 0.63}
\definecolor{lasallegreen}{rgb}{0.03, 0.47, 0.19}
\definecolor{hanpurple}{rgb}{0.32, 0.09, 0.98}
\definecolor{green(pigment)}{rgb}{0.0, 0.65, 0.31}
\usepackage[pagebackref=true,breaklinks=true,letterpaper=true,colorlinks,bookmarks=false]{hyperref}
\hypersetup{colorlinks,linkcolor={red},citecolor={hollywoodcerise},urlcolor={red}}  
\usepackage{wrapfig} 
\usepackage[font=small]{caption} 

\definecolor{cvprblue}{rgb}{0.21,0.49,0.74}
\def\paperID{948} % *** Enter the Paper ID here
\def\confName{CVPR}
\def\confYear{2024}

\title{Semantics, Distortion, and Style Matter: Towards Source-free UDA for Panoramic Segmentation}
\author{Xu Zheng$^{1}$ \quad Pengyuan Zhou$^{3}$ \quad Athanasios V. Vasilakos$^{4}$ \quad Lin Wang$^{1}$$^{,2}$\thanks{Corresponding author.}\\
$^{1}$AI Thrust, HKUST(GZ)  \quad $^{2}$Dept. of CSE, HKUST \quad $^{3}$	Aarhus University \quad $^{4}$ University of Agder\\
{\tt\small zhengxu128@gmail.com, pengyuan.zhou@ece.au.dk,  th.vasilakos@gmail.com, linwang@ust.hk}
\\
\small{Project Page: \url{https://vlislab22.github.io/360SFUDA/}}
}

\begin{document}
\maketitle
\begin{abstract}
This paper addresses an interesting yet challenging problem-- source-free unsupervised domain adaptation (SFUDA) for pinhole-to-panoramic semantic segmentation--given only a pinhole image-trained model (\ie, source) and unlabeled panoramic images (\ie, target). Tackling this problem is nontrivial due to the semantic mismatches, style discrepancies, and inevitable distortion of panoramic images. To this end, we propose a novel method that utilizes Tangent Projection (TP) as it has less distortion and meanwhile slits the equirectangular projection (ERP) with a fixed FoV to mimic the pinhole images. Both projections are shown effective in extracting knowledge from the source model. However, the distinct projection discrepancies between source and target domains impede the direct knowledge transfer; thus, we propose a panoramic prototype adaptation module (PPAM) to integrate panoramic prototypes from the extracted knowledge for adaptation. We then impose the loss constraints on both predictions and prototypes and propose a cross-dual attention module (CDAM) at the feature level to better align the spatial and channel characteristics across the domains and projections. Both knowledge extraction and transfer processes are synchronously updated to reach the best performance. Extensive experiments on the synthetic and real-world benchmarks, including outdoor and indoor scenarios, demonstrate that our method achieves significantly better performance than prior SFUDA methods for pinhole-to-panoramic adaptation. 
\end{abstract}

 \begin{figure}[t!]
    \centering
    \includegraphics[width=\linewidth]{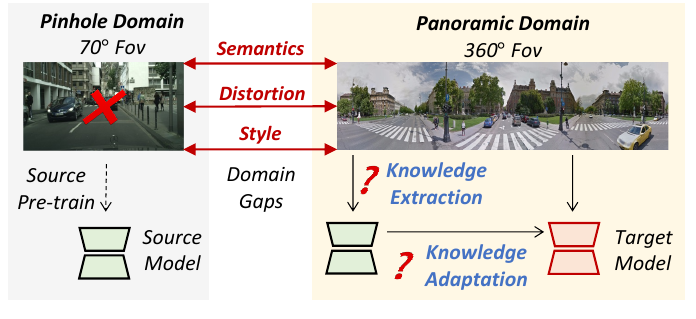}
    % \vspace{-24pt}
    \caption{We address a new problem of achieving source-free pinhole-to-panoramic adaptation for segmentation.}
    %(b) Visual comparison between SFDA~\cite{liu2021source} and our method.}
    \label{fig:cover}
    % \vspace{-16pt}
\end{figure}

% \vspace{-15pt}
\section{Introduction}
\label{Sec:intro}
The comprehensive scene perception abilities of $360^\circ$ cameras have made them highly popular for applications, such as autonomous driving~\cite{360survey}. In contrast to pinhole cameras that capture 2D planer images with a limited field-of-view (FoV), $360^\circ$ cameras offer a much wider FoV of $360^\circ \times 180^\circ$. As a result, research on panoramic semantic segmentation~\cite{yang2019pass, yang2020omnisupervised, zhang2022bending, p2pda, PCS} has been actively explored to achieve dense scene understanding for intelligent systems. 

Generally, the spherical data captured by the $360^\circ$ cameras is always projected into 2D planar representations, \eg, Equirectangular Projection (ERP), to be aligned with the existing imaging pipeline~\cite{360survey} while preserving the omnidirectional information \footnote{In this paper, omnidirectional and panoramic images are interchangeably used, and ERP images often indicate panoramic images.}. However, ERP suffers from the inevitable distortion and object deformation due to the non-uniformly distributed pixels~\cite{zheng2023both}. Meanwhile, learning effective panoramic segmentation models is often impeded by the lack of large precisely labeled datasets due to the difficulty of annotation. For these reasons, some unsupervised domain adaptation (UDA) methods~\cite{zhang2022bending, zhang2022behind, zheng2023both} have been proposed to transfer the knowledge from the pinhole image domain to the panoramic image domain. In some crucial application scenarios, \eg, autonomous driving, source datasets are not always accessible due to privacy and commercial issues, such as data portability and transmission costs. One typical example is the recent large model, SAM~\cite{SAM}, which brings significant progress in instance segmentation for pinhole images; however, the source datasets are too large (10TB) to be reused in end-tasks, such as \cite{kirillov2023segment}.

 \textbf{Motivation:} In this paper, we probe an interesting yet challenging problem: \textit{source-free UDA (SFUDA) for panoramic segmentation, in which only the source model (pretrained with pinhole images) and unlabeled panoramic images are available}. As shown in Fig.~\ref{fig:cover} \textcolor{red}{(a)}, different from existing SFUDA methods,~\eg, \cite{liu2021source,ye2021source,yang2022source} for the pinhole-to-pinhole image adaptation, transferring knowledge from the pinhole-to-panoramic image domain is hampered by:\textbf{ 1)} semantic mismatch caused by the different FoV between the pinhole and $360^\circ$ cameras, \ie, 70$^\circ$ vs. 360$^\circ$; 
\textbf{2)} inevitable distortion of the ERP; \textbf{3)} style discrepancies caused by the distinct camera sensors and captured scenes. In Tab.~\ref{Tab:City2PASS}, we show that naively adapting existing SFUDA methods to our problem leads to a limited performance boost. 
%(also see Fig.~\ref{fig:cover}(b)). 
% This makes it a challenging problem regarding how to effectively extract knowledge from the source model with only panoramic images.

\textbf{Contributions:} To this end, we propose a novel SFUDA method that effectively extracts knowledge from the source model with only panoramic images and transfers the knowledge to the target panoramic domain.
%[add key insight here!]
\textit{Our key idea is to leverage the multi-projection versatility of 360$^\circ$ data for efficient domain knowledge transfer}.
Our method enjoys two key technical contributions. 
Specifically, we use Tangent Projection (TP) and divide the ERP images into patches with a fixed FoV, dubbed Fixed FoV Projection (FFP), to extract knowledge from the source model with less distortion and similar FoV to the pinhole images. 
% Specifically, we first utilize Tangent Projection (TP) as it has less distortion than ERP and meanwhile divide the ERP images into patches with a fixed FoV, dubbed Fixed FoV Projection (FFP). Intuitively, the FFP patches have a similar FoV to the source pinhole images, while the TP images have less distortion than the ERP images. 
Both projections make it possible to effectively extract knowledge from the source model. 
However, directly transferring the extracted knowledge to the target model is hardly approachable due to the distinct projection gaps. Thus, we propose a panoramic prototype adaptation module (PPAM) to obtain \textit{class-wise semantic prototypes} from the features and predictions of the source model with TP and FFP images (Sec.~\ref{Sec.3.2}). Then, these prototypes are integrated together to obtain the global panoramic prototypes for knowledge adaptation, which is updated across the adaptation procedure. 
Moreover, our proposed PPAM also fine-tunes the source model to promote better knowledge extraction using prototypes extracted from FFP images. Aligning the prototypes from each FFP image enables the source model to become more aware of distortion and semantics across the FoV.

%To transfer the extracted knowledge to the unlabeled target panoramic image domain, we first impose the prediction-level and prototype-level loss constraints. 
We initially apply both prediction-level and prototype-level loss constraints to facilitate knowledge transfer to the unlabeled target panoramic domain. 
Concretely, the FFP predictions of the source model are rebuilt together to provide a pseudo-supervision signal for the target model. The prototype-level loss constraint is performed between the panoramic prototypes from PPAM and the prototypes from the target model's features and predictions on the ERP images.
Moreover, knowledge from the source model is not limited to predictions and prototypes, high-level features also contain crucial image characteristics that can enhance the performance of the target model. Consequently, we propose a Cross-Dual Attention Module (\textbf{CDAM}) that \textit{aligns spatial and channel characteristics between domains} to fully utilize the knowledge from the source model and address the style discrepancy problem (Sec.~\ref{Sec:3.3}). Specifically, CDAM reconstructs the source model features from FFP images to provide a panoramic perception of the surrounding environment and aligns them with the ERP features from the target model for effective knowledge transfer.

We conduct extensive experiments on both synthetic and real-world benchmarks, including outdoor and indoor scenarios.
% Different from prior methods, we solve a new problem by transferring knowledge from pinhole pre-trained models to panoramic images without accessing the source data. 
As no directly comparable works exist, we adapt the state-of-the-art (SoTA) SFUDA methods~\cite{liu2021source, zhang2021prototypical, kundu2021generalize, huang2021model, yang2022source, guo2022simt} -- designed for pinhole-to-pinhole image adaptation -- to our problem in addressing the panoramic semantic segmentation.
The results show that our framework significantly outperforms these methods by large margins of +6.37\%, +11.47\%, and +10.99\% on three benchmarks. 
We also evaluate our method against UDA methods~\cite{zhang2022behind, zhang2022bending, zheng2023both, zheng2023look}, using the source pinhole image, the results demonstrate its comparable performance.
%Also, we compare with the UDA methods for panoramic segmentation~\cite{zhang2022behind, zhang2022bending, zheng2023both, zheng2023look}, using the source pinhole images. The results show that our method is on par with these methods.

% In summary, our main contributions are as follows: \textbf{(I)} We address a new problem by transferring semantic knowledge from models pre-trained in the pinhole domain to the panoramic domain without access to the pinhole images. \textbf{(II)} We propose an end-to-end framework that incorporates PPAM and CDAM modules to achieve prototype, feature, and prediction-level knowledge adaptation. \textbf{ (III)} We conduct intensive experiments on both synthetic and real-world knowledge transfer scenarios, and experimental results demonstrate the effectiveness of our proposed SFUDA framework.

% \vspace{-6pt}
\section{Related Work}
\subsection{Source-free UDA for Segmentation}
% The UDA problem has received significant attention as a solution to  
UDA aims to mitigate the impact of domain shift caused by data distribution discrepancies in downstream computer vision tasks, such as semantic segmentation~\cite{zhang2019category,chen2019domain,hoyer2022daformer,vu2019dada,zou2018unsupervised,stan2021unsupervised,fleuret2021uncertainty,shen2021unsupervised,vu2019advent,pan2020unsupervised,araslanov2021self, zheng2022uncertainty, chen2022uncertainty, zhu2023good, zheng2022transformer, chen4617170frozen, zheng2023distilling, chen2023clip, xie2023adversarial}. However, the source domain data may not always be accessible due to the privacy protection and data storage concerns. Intuitively, source-free UDA (SFUDA)~\cite{kundu2021generalize,yeh2021sofa,huang2021model} methods are proposed to adapt source models to a target domain without access to the source data. Existing SFUDA methods for semantic segmentation primarily focus on source data estimation~\cite{ye2021source,yang2022source} or self-training~\cite{kundu2021generalize,zhao2022source,liu2021source,bateson2022source} for pinhole images. \textit{In this paper, we make the first attempt at achieving SFUDA from the pinhole image domain to the panoramic domain}. This task is nontrivial to be tackled due to the semantic mismatches, style discrepancies, and inevitable distortion of panoramic images. Unlike these methods that focus on the source domain data estimation~\cite{liu2021source, ye2021source}, we propose a novel SFUDA method that effectively extracts knowledge from the source model with only panoramic images and transfers the knowledge to the target panoramic image domain. Experiments also show that naively applying these methods leads to less optimal performance (See Tab.~\ref{Tab:City2PASS}).

% Unlike these methods that focus on the source domain data estimation~\cite{liu2021source, ye2021source}, we focus on extracting knowledge purely from the source model and take full advantage of the different projection types of the spherical data for better knowledge extraction. 
\begin{figure*}[h!]
    \centering
    \includegraphics[width=0.9\textwidth]{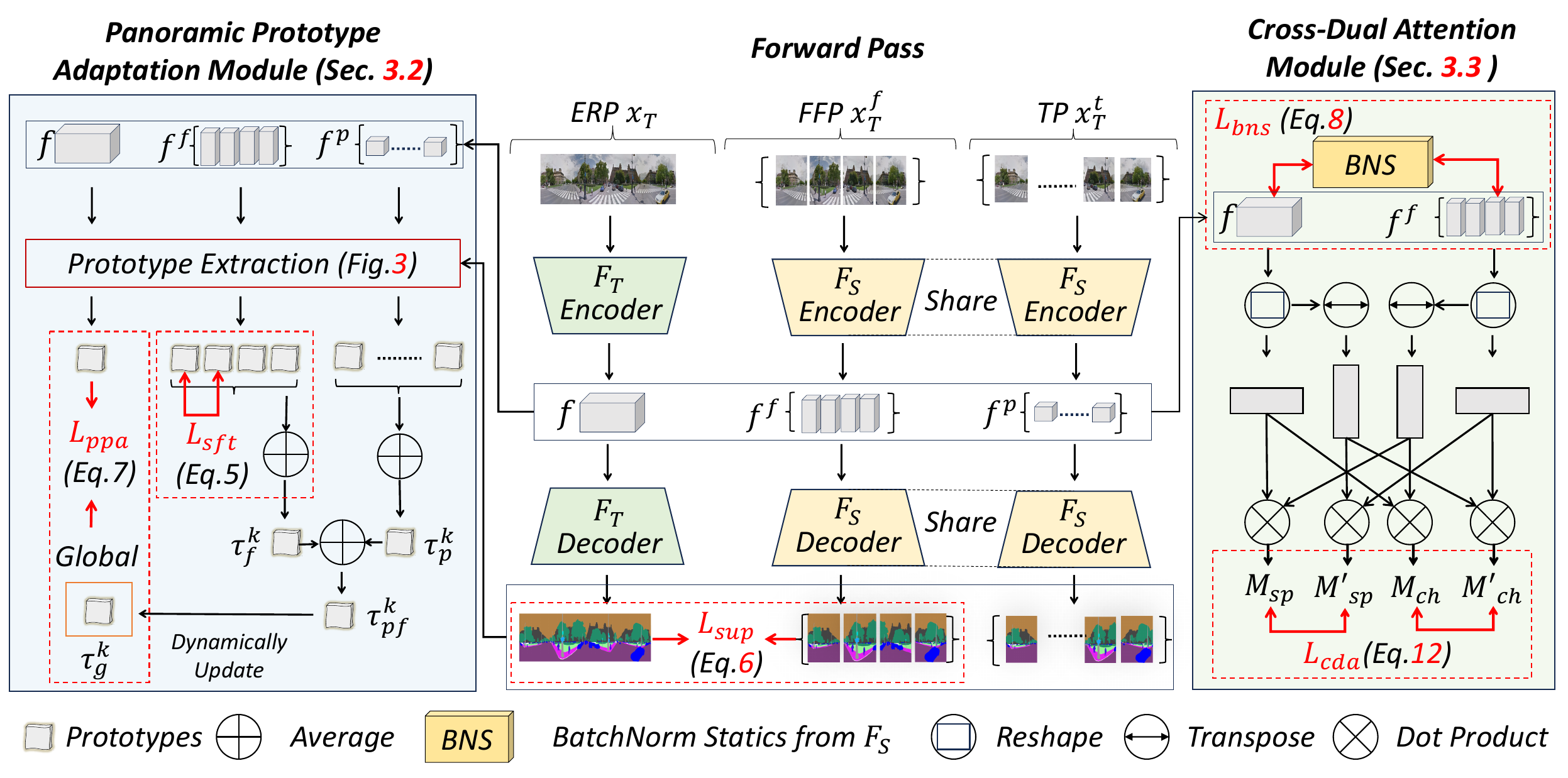}
    % \vspace{-12pt}
    \caption{Overall framework of our proposed SFUDA for panoramic semantic segmentation. %(a) Illustration of cross dual attention module (CDAM), (b) forward passes of the source and target models; (c) illustration of the panoramic prototype adaptation module (PPAM).
    }
    \label{fig:framework}
    % \vspace{-12pt}
\end{figure*}

\subsection{UDA for Panoramic Semantic Segmentation}
% UDA for panoramic semantic segmentation 
It can be classified into three types, including adversarial training~\cite{Hoffman2018CyCADACA,Choi2019SelfEnsemblingWG,Sankaranarayanan2018LearningFS,Tsai2018LearningTA,zheng2023both}, pseudo labeling~\cite{Liu2021PanoSfMLearnerSM,Zhang2021DeepPanoContextP3,Wang2021DomainAS,Zhang2017CurriculumDA} and prototypical adaptation methods~\cite{zhang2022bending,zhang2022behind}. Specifically, the first line of research applies alignment approaches to capture the domain invariant characteristics of images~\cite{Hoffman2018CyCADACA,Li2019BidirectionalLF,Murez2018ImageTI}, feature~\cite{Hoffman2018CyCADACA,Chen2019ProgressiveFA,Hoffman2016FCNsIT,zheng2023both} and predictions~\cite{Luo2019TakingAC,MelasKyriazi2021PixMatchUD}. The second type of methods generates pseudo labels for the target domain training. The last line of research, \eg, Mutual Prototype Adaption (MPA)~\cite{zhang2022bending}, mutually aligns the high-level features with the prototypes between domain. However, these methods treat panoramic images as pinhole images when extracting prototypes, ignoring the intricate semantic, object correspondence, and distortion information brought by the panoramic FoV. \textit{We are the first to address the SFUDA problem for panoramic segmentation. Considering the distinct projection discrepancies between source and target domains, we propose a PPAM to integrate the global panoramic prototypes from the extracted knowledge for adaptation.}

\section{Methodology}   
%% \vspace{-8pt}
\subsection{Overview}
%We describe the proposed SFUDA framework for panoramic segmentation, as shown in Fig.~\ref{fig:framework}. 
The overall framework for panoramic segmentation is shown in Fig.~\ref{fig:framework}.
With only the source model $F_S$ available and given the unlabeled panoramic image data $D_T$, we aim to train a target model $F_T$ that adapts knowledge from $F_S$ to the common $K$ categories across both domains.

%Given the target domain of unlabeled panoramic image data $D_T$, and in the absence of both source domain data and labels, we have access only the source model $F_S$. Assuming a commonality of $K$ categories across both domains, our objective is to train a target model $F_T$, facilitating the adaptation of knowledge $F_S$ to learn from the unlabeled panoramic images.
%Assume that we are given the target domain, \ie, panoramic image data $D_T$ \textit{without labels}. However, the source domain, \ie, pinhole image data and labels are both \textit{unavailable}, and only the source model $\mathcal{F}_S$ is given. We assume both domains share $K$ common categories to achieve panoramic segmentation. Our goal is to train a target model $\mathcal{F}_T$ learning unlabeled panoramic images by adapting knowledge from the source model $\mathcal{F}_S$. 
% We propose an end-to-end SFUDA framework that addresses the challenges of transferring knowledge from the pinhole-to-panoramic image domain. 

Unlike the pinhole image-to-image adaptation~\cite{liu2021source,ye2021source,yang2022source}, pinhole-to-panoramic image domain adaptation is hampered by three key factors,
specifically: semantic mismatch due to FoV variations (70$^\circ$ vs. 360$^\circ$), inevitable distortion in ERP, and ubiquitous style discrepancies in unsupervised domain adaptation (UDA) (refer to Fig.\ref{fig:cover}~\textcolor{red}{(a)}).
%namely \textbf{semantic mismatch} caused by the FoV discrepancies across domains, \textbf{inevitable distortion} of ERP, and \textbf{style discrepancies} which always exist in UDA problems (See Fig.~\ref{fig:cover}(a)). 
Therefore, naively applying existing SFUDA methods exhibits sub-optimal segmentation performance (See Tab.~\ref{Tab:City2PASS}), while UDA methods with source data, \eg, \cite{liu2021source} for panoramic segmentation do not account for the semantic mismatch between the pinhole and panoramic images. Intuitively, the key challenges are
: \textbf{1)} how to extract knowledge from the source model with only panoramic images and \textbf{2)} how to transfer knowledge to the target panoramic image domain.

\noindent \textbf{Our key idea} \textit{is to leverage the multi-projection versatility of 360$^\circ$ data for efficient domain knowledge transfer.}

Concretely, to address the first challenge (Sec.~\ref{Sec.3.2}), we use the Tangent Projection (TP)
which is characterized by a reduced distortion issue compared to the ERP images~\cite{eder2020tangent} to extract knowledge from the source model.
%for its capacity to maintain high accuracy proximal to the tangency point, thereby rendering it particularly suited for minimizing distortion~\cite{eder2020tangent}.
Concurrently, ERP images are segmented into discrete patches, each possessing a constant FoV to mimic the pinhole images, dubbed Fixed FoV Projection (FFP).
Both projections make it possible to effectively extract knowledge from the source model. 
%As shown in Fig.~\ref{fig:framework}, t
The distinct projection formats make it impossible to directly transfer knowledge between domains, thus we propose a Panoramic Prototype Adaptation Module (PPAM) to obtain panoramic prototypes for adaptation. 
To address the second challenge (Sec.~\ref{Sec:3.3}), we first impose prediction and prototype level loss constraints, and propose a Cross-Dual Attention Module (CDAM) at the feature level to transfer knowledge and further address the style discrepancies.

\subsection{Knowledge Extraction}
\label{Sec.3.2}
As depicted in Fig.~\ref{fig:framework}, given the target domain (\ie, panoramic domain) ERP images $D_T=\{x_T|x_T \in \mathbf{R}^{H\times W\times 3}\}$, we first project them into TP images $D_T^{t}=\{x_T^{t}|x_T^{t} \in \mathbf{R}^{h\times w\times 3}\}$ and FFP images $D_T^{f}=\{x_T^{f}|x_T^{f} \in \mathbf{R}^{H\times W/4\times 3}\}$ for effectively extracting knowledge from the source model. Note that one ERP image corresponds to $18$ TP images as ~\cite{li2022omnifusion, zheng2023both} and $4$ FFP images with a fixed FoV of $90^\circ$ (See Sec.~\ref{Sec:ab}).
To obtain the features and predictions from the source model for knowledge adaptation, the two types of projected images are first fed into the source model with batch sampling:
\begin{equation}
\setlength{\abovedisplayskip}{3pt}
\setlength{\belowdisplayskip}{3pt}
    P^p, f^p =F_S(x_T^{t}), \qquad P^f, f^f =F_S(x_T^{f}),
\end{equation}
% \begin{equation}
%    \{P^p = \sum_{a=1}^{18}P^p_a\}, \{F^p = \sum_{a=1}^{18}F^p_a\} = \mathcal{F}_S(x_T^{t}), \qquad 
%    \{P^f = \sum_{b=1}^{4}P^f_b\}, \{F^f = \sum_{b=1}^{4}F^f_b\} = \mathcal{F}_S(x_T^{f}),
% \end{equation}
where $f^p$, $f^f$, $P^p$, and $P^f$ are the source model features and predictions of the input TP and FFP images, respectively. 
For the target panoramic images, $x_T$ is fed into $F_T$ to obtain the target model features $f$ and predictions $P$ of the input batch of ERP images as $P, f = F_T(x_T)$.
% \begin{equation}
%     P, F = \mathcal{F}_T(x_T),
% \end{equation}
% where $F$ and $P$ are the target model features and predictions of the input batch of ERP images. 
However, the distinct projection formats of the input data in the source and target models make it difficult to align their features directly, thus we propose a Panoramic Prototype Adaptation Module (PPAM) to obtain panoramic prototypes for adaptation.  
\begin{figure}[t!]
    \centering
    \includegraphics[width=\linewidth]{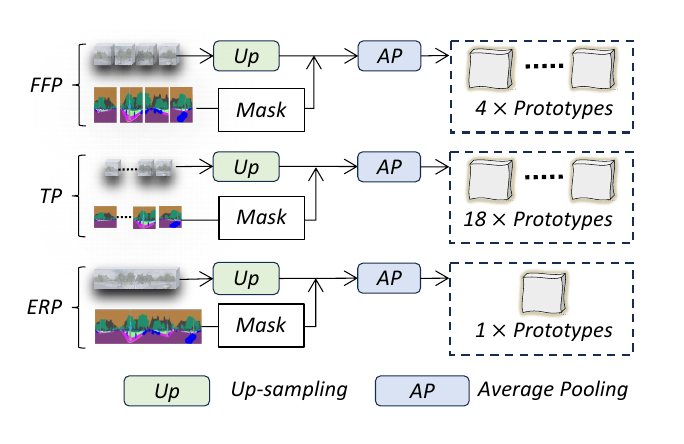}
    % \vspace{-28pt}
    \caption{Illustration of the prototype extraction (PE) in the panoramic prototype adaptation module (PPAM).
    }
    \label{fig:protoytpe_extraction}
    % \vspace{-16pt}
\end{figure}

\noindent \textbf{Panoramic Prototype Adaptation Module (PPAM)}
Compared to prior UDA methods using prototypical adaptation, \eg, MPA~\cite{zhang2022bending, zhang2022behind}, our PPAM possesses three distinct characteristics: \textbf{(a)} class-wise prototypes are obtained from TP and FFP images to alleviate distortion and semantic mismatch problems; \textbf{(b)} global prototypes are iteratively updated with prototypes from two projections during the whole training procedure; \textbf{(c)} hard pseudo-labels are softened in the high-level feature space to obtain prototypes with different projection of panoramic images, indicating that the knowledge from the source model is fully utilized. 

%for panoramic semantic segmentation.
% To convert the source model predictions \( P^p \) and \( P^f \) into pseudo labels for knowledge transfer, we define \(\hat{y}^p_{(h,w,k)}\) as 1 when \( k \) equals the index at which \( P^p_{h,w,:} \) attains its maximum value, \ie, \(\hat{y}^p_{(h,w,k)} = 1_{k \doteq \arg\max(P^p_{h,w,:}) }\). Similarly, \(\hat{y}^f_{(H,W/4,k)}\) is set to 1 where \( k \) matches the index of the maximal value in \( P^f_{H,W/4,:} \), expressed as \(\hat{y}^f_{(H,W/4,k)} = 1_{k \doteq \arg\max(P^f_{H,W/4,:}) }\). Here, \( k \) denotes the semantic category.
Specifically, we project the source model predictions $P^p$, $P^f$ into pseudo labels with $\hat{y}^p_{(h,w,k)} = 1_{k \doteq argmax(P^p_{h,w,:}) }$ and $\hat{y}^f_{(H,W/4,k)} = 1_{k \doteq argmax(P^f_{H,W/4,:}) }$, respectively. Here, $k$ denotes the semantic category.
\begin{align}
\setlength{\abovedisplayskip}{3pt}
\setlength{\belowdisplayskip}{3pt}
    &\hat{y}^p_{(h,w,k)} = 1_{k \doteq argmax(P^p_{h,w,:}) }, \qquad \notag \\
    &\hat{y}^f_{(H,W/4,k)} = 1_{k \doteq argmax(P^f_{H,W/4,:}) }.
\end{align}
Subsequently, we obtain the class-specific masked features by integrating the up-sampled features with the corresponding pseudo labels \(\hat{y}^p_{(h,w,k)}\) and \(\hat{y}^f_{(H,W/4,k)}\).
Notably, the prototypes $\sum_{a=1}^{18} (\tau_p^k)_a$ and $\sum_{b=1}^4 (\tau_f^k)_b$ for TP and FFP images are obtained by masked average pooling (MAP) operation, as shown in Fig.~\ref{fig:protoytpe_extraction}. Within each projection, PPAM first integrates the prototypes:
\begin{equation}
\setlength{\abovedisplayskip}{3pt}
\setlength{\belowdisplayskip}{3pt}
    \tau_p^k = avg(\sum_{a=1}^{18} (\tau_p^k)_a), \qquad 
    \tau_f^k = avg(\sum_{b=1}^4 (\tau_f^k)_b).
\end{equation}
As shown in Fig.~\ref{fig:framework}, $\tau_p^k$ and $\tau_f^k$ are integrated together as $\tau_{pf}^k$ to preserve the less distortion characteristics of $\tau_p^k$ and the similar scale semantics of $\tau_f^k$. The $\tau_{pf}^k$ is then used to update the panoramic global prototype $\tau_g^k$, which is iteratively updated with $\tau_{pf}^k$. To obtain more accurate and reliable prototypes, we update $\tau_g^k$ and $\tau_{pf}^k$ as follows:
\begin{equation} 
\label{Eq:iter}
\setlength{\abovedisplayskip}{3pt}
\setlength{\belowdisplayskip}{3pt}
    \tau_g^i = \frac{1}{i}(\tau_{pf}^k)^i + (1 - \frac{1}{i})(\tau_g^k)^{i-1},
\end{equation}
where $(\tau_g^k)^i$ and $(\tau_{pf}^k)^i$ are the prototypes for category $k$ in the $i-$th training epoch, $(\tau_g^k)^{i-1}$ is the panoramic global prototype saved in the last training epoch, $i$ is the current epoch number.
The panoramic global prototype $\tau_g^k$ is then used to give supervision for the target prototype $\tau_t^k$ obtained from $P$ and $f$ with the same operations.

Besides extracting prototype knowledge from the source model, PPAM also fine-tunes the source model to improve the effectiveness of knowledge extraction. Specifically, since each ERP image can be projected to 4 FFP images, the source model's extracted features $f_f$ have 4 pieces of FFP features. As the content of all the features is within the same ERP image, we propose to align the class-wise prototypes from each piece of the features in PPAM to enhance the model's performance. Concretely, the prototypes $\sum_{\alpha=1}^4 \tau_{\alpha}$ of the four FFP features are obtained through the same operations with $\tau_g^t$. 
Each FFP image captures a non-overlapping $90^\circ$ FoV, resulting in distinct distortions, and similar content in each FFP image. Aligning the prototypes from each FFP image enhances distortion-awareness ability in the source model and helps to explore complementary semantic content in each FFP image.
The MSE loss is imposed between each two of the prototypes as follows:
\begin{equation}
\setlength{\abovedisplayskip}{3pt}
\setlength{\belowdisplayskip}{3pt}
    \mathcal{L}_{sft} = \sum_{\alpha \neq \beta}^4\{\frac{1}{K}\sum_{k \in K}((\tau_f^k)_{\alpha}-(\tau_f^k)_{\beta})^2\}.
\end{equation}
Note that $\mathcal{L}_{sft}$ is used to fine-tune the source model $F_S$.

\begin{table*}[h!]
\setlength{\tabcolsep}{4pt}
\resizebox{0.99\textwidth}{!}{
\begin{tabular}{l|c|c|ccccccccccccc|c}
\toprule
Method & SF & mIoU & Road & S.W. & Build. & Wall & Fence & Pole & Tr.L. & Tr.S. & Veget. & Terr. & Sky & Pers. & Car & $\Delta$ \\ \midrule
PVT~\cite{wang2021pyramid} SSL & \XSolidBrush & 38.74 & 55.39 & 36.87 & 80.84 & 19.72 & 15.18 & 8.04 & 5.39 & 2.17 & 72.91 & 32.01 & 90.81 & 26.76 & 57.40 &-\\
PVT~\cite{wang2021pyramid} MPA & \XSolidBrush & 40.90 & 70.78 & 42.47 & 82.13 & 22.79 & 10.74 & 13.54 & 1.27 & 0.30 & 71.15 & 33.03 & 89.69 & 29.07 & 64.73 &-\\
% Trans4PASS~\cite{zhang2022behind} SSL & \XSolidBrush & 43.17 & 73.72 & 43.31 & 79.88 & 19.29 & 16.07 & 20.02 & 8.83 & 1.72 & 67.84 & 31.06 & 86.05 & 44.77 & 68.58 &-\\
% Trans4PASS~\cite{zhang2022behind} MPA & \XSolidBrush & 45.29 & 67.28 & 43.48 & 83.18 & 22.02 & 21.98 & 22.72 & 7.86 & 1.52 & 73.12 & 40.65 & 91.36 & 42.69 & 70.87 &-\\ 
\midrule
Source w/ seg-b1 & \Checkmark & 35.81 & 63.36 & 24.09 & 80.13 & \textbf{15.68} & 13.39 & 16.26 & 7.42 & 0.09 & 62.45 & 20.20 & 86.05 & 23.02 & 53.37 &- \\
SFDA w/ seg-b1~\cite{liu2021source} & \Checkmark & 38.21 & 68.78 & 30.71 & 80.37 & 5.26 & 18.95 & 20.90 & 5.25 & 2.36 & 70.19 & 23.30 & \underline{90.20} & 22.55 & 57.90 &+2.40 \\
ProDA w/ seg-b1~\cite{zhang2021prototypical} & \Checkmark & 37.37 & 68.93 & 30.88 & 80.07 & 4.17 & 18.60 & 19.72 & 1.77 & 1.56 & 70.05 & 22.73 & \textbf{90.60} & 19.71 & 57.04 & +2.73 \\
GTA w/ seg-b1~\cite{kundu2021generalize} & \Checkmark & 36.00& 64.61 & 20.04 & 79.04 & 8.06 & 15.36 & 19.86 & 6.02 & 2.13 & 65.77 & 17.75 & 84.56 & 26.71 & 58.13 & +0.19 \\
HCL w/ seg-b1~\cite{huang2021model} & \Checkmark & 38.38 & 68.82 & 30.41 & 80.37 & 5.88 & 20.18 & 20.10 & 4.23 & 2.11 & 70.50 & 24.74 & 89.89 & 22.65 & 59.04 & +2.57 \\
DATC w/ seg-b1~\cite{yang2022source} & \Checkmark & 38.54 & 69.48 & 26.96 & 80.68 & 11.64 & 15.24 & 20.10 & \textbf{9.33} & 0.55 & 66.11 & 24.31 & 85.16 & 30.90 & 60.58 &+2.73\\ 
Simt w/ seg-b1~\cite{guo2022simt} & \Checkmark & 37.94 & 68.47 & 29.51 & 79.62 & 6.78 & 19.20 & 19.48 & 2.31 & 1.33 & 68.85 & \underline{26.55} & 89.30 & 22.35 & 59.49 & +2.13 \\ \midrule
Ours w/ seg-b1 & \Checkmark & \underline{41.78} & \textbf{70.17} & \textbf{33.24} & \textbf{81.66} & \underline{13.06} & \underline{23.40} & \underline{23.37} & 7.63 & \underline{3.59} & \underline{71.04} & 25.46 & 89.33 & \underline{36.60} & \underline{64.60} & \underline{+5.97}\\ 
Ours w/ seg-b2 & \Checkmark & \textbf{42.18} & \underline{69.99} & \underline{32.28} & \underline{81.34} & 10.62 & \textbf{24.35} & \textbf{24.29} & \underline{9.19} & \textbf{3.63} & \textbf{71.28} & \textbf{30.04} & 88.75 & \textbf{37.49} & \textbf{65.05} & \textbf{+6.37}\\ 
\bottomrule
\end{tabular}}
% \vspace{-8pt}
\caption{Experimental results on the S-to-D scenario, the overlapped 13 classes of two datasets are used to test the UDA performance. The \textbf{bold} and \underline{underline} denote the best and the second-best performance in source-free UDA methods, respectively.}
% \vspace{-12pt}
\label{Tab:syn2dense}
\end{table*}
\subsection{Knowledge Adaptation}
\label{Sec:3.3}
To adapt knowledge to the target domain, we impose the loss constraints on both predictions and prototypes and propose a cross-dual attention module (\textbf{CDAM}) at the feature level to better align the spatial and channel characteristics across the domains and projections. 
Specifically, the predictions of the FFP patch images
are stitched to reconstruct an ERP image. The ERP image is then passed to the source model $F_S$ to predict a pseudo label, which serves as the supervision for the ERP predictions of the target model $F_T$. For simplicity, we use the Cross-Entropy (CE) loss, which is formulated as:
\begin{equation}
    \mathcal{L}_{sup} = CE(P, 1_{k \doteq argmax(\{Rebuild(P^f_{H,W/4,:})\}) }).
\end{equation}
And the prototype-level knowledge transfer loss is achieved by Mean Squared Error (MSE) loss between the panoramic global prototype $\tau_g^k$ and the target prototype $\tau_t^k$ :
\begin{equation}
    \mathcal{L}_{ppa} = \frac{1}{K}\sum_{k \in K}(\tau_g^k-\tau_t^k)^2.
\end{equation}
With loss $\mathcal{L}_{ppa}$, the prototypes are pushed together to transfer the source-extracted knowledge to the target domain.
In summary, with the proposed PPAM, we effectively address the distortion and semantic mismatch problems at the prediction and prototype level, we now tackle the style discrepancy problem at the feature level.
\begin{figure*}[h!]
    \centering\includegraphics[width=\textwidth]{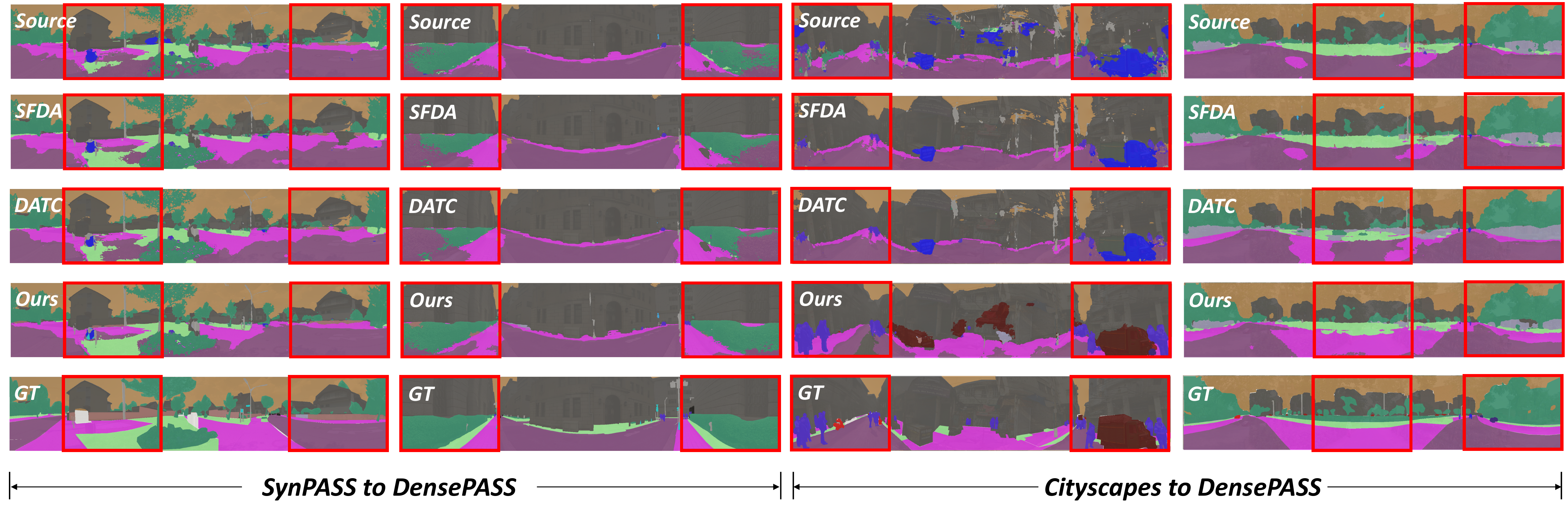}
    % \vspace{-24pt}
    \caption{Example visualization results. (a) source, (b) SFDA~\cite{liu2021source}, (c) DATC~\cite{yang2022source}, (d) Ours, (e) Ground Truth (GT).}
    % \vspace{-6pt}
    \label{fig:visualization}
    %% \vspace{-15pt}
\end{figure*}
\begin{table*}[h!]
\centering
\setlength{\tabcolsep}{10pt}
\resizebox{0.99\textwidth}{!}{
\begin{tabular}{lcccccccccc|c}
\toprule
Method & SF & mIoU & Person & Rider & Car & Truck & Bus & Train & Motor & Bike &$\Delta$ \\ \midrule
Trans4PASS-T~\cite{zhang2022bending}  & \XSolidBrush & 53.18 & 48.54 & 16.91 & 79.58 & 65.33 & 55.76 & 84.63 & 59.05 & 37.61 & - \\ 
Trans4PASS-S~\cite{zhang2022bending} & \XSolidBrush & 55.22 & 48.85 & 23.36 & 81.02 & 67.31 & 69.53 & 86.13 & 60.85 & 39.09 & - \\ 
DAFormer~\cite{hoyer2022daformer} & \XSolidBrush & 54.67 & 49.69 & 25.15 & 77.70 & 63.06 & 65.61 & 86.68 & 65.12 & 48.13 & - \\ 
DPPASS~\cite{zheng2023both} & \XSolidBrush & 55.30 & 52.09 & 29.40 & 79.19 & 58.73 & 47.24 & 86.48 & 66.60 & 38.11 & - \\
DATR~\cite{zheng2023look} & \XSolidBrush & 56.81 & 54.62 & 29.50 & 80.03 & 67.35 & 63.75 & 87.67 & 67.57 & 37.10 & -\\
\midrule
Source w/ seg-b1 & \Checkmark & 38.65 & 40.93 & 10.89 & 67.67 & 36.86 & 15.56 & 26.43 & 42.68 & 27.16 & - \\ 
SFDA w/ seg-b1~\cite{liu2021source}& \Checkmark & 42.70 & 41.65 & 8.46 & 69.97 & 47.48 & 33.24 & 72.01 & 47.61 & 32.77 & +4.05\\  
DTAC w/ seg-b1~\cite{yang2022source} & \Checkmark & 43.06 & 43.51 & 8.35 & 70.10 & 35.79 & 40.73 & 70.52 & 49.49 & 32.94 & +4.41 \\ \midrule
Ours w/ seg-b1 & \Checkmark & \underline{48.78} & \underline{45.36} & \underline{15.83} & \underline{75.70} & \textbf{49.16} & \underline{55.68} & \textbf{82.07} & \underline{54.82} & \underline{33.76} & \underline{\textbf{+10.13}} \\ 
Ours w/ seg-b2 & \Checkmark & \textbf{50.12} & \textbf{49.92} & \textbf{27.22} & \textbf{76.22} & \underline{47.81} & \textbf{64.13} & \underline{79.47} & \textbf{56.83} & \textbf{35.76} & \textbf{+11.47}\\ 
\bottomrule
\end{tabular}}
% \vspace{-8pt}
\caption{Experimental results of 8 selected categories in panoramic semantic segmentation on C-to-D. SF: Source-free UDA. The \textbf{bold} and \underline{underline} denote the best and the second-best performance in source-free UDA methods, respectively.}
% \vspace{-10pt}
\label{Tab:City2PASS}
\end{table*}

\noindent \textbf{Cross Dual Attention Module (CDAM).}
%As semantic information and contextual relationships within an panoramic image performs a significant effect on segmentation, thus CDAM aims to transfer the contextual correspondence within an ERP captured by the source model.
Inspired by the dual attention, focusing on spatial and channel characteristics~\cite{liu2021source}, our CDAM imitates the spatial and channel-wise distributions of features to alleviate the style discrepancies. 
Different from \cite{liu2021source} suggesting to minimize the distribution distance of the dual attention maps between the fake source (FFP images) and target data (ERP images), our CDAM focuses on \textit{aligning the distribution between FFP and ERP of the panoramic images} rather than introducing additional parameters and computation cost in estimating source data. As shown in Fig.~\ref{fig:framework}, we reconstruct the FFP features $F^f$ to ensure that the rebuilt feature $F'$ has the same spatial size as $F$. Before the cross dual attention operation, we apply a Batch Normalization Statics (BNS) guided constraint on $F$ and $F'$. Since the BNS of the source model should satisfy the feature distribution of the source data, we align $F$ and $F'$ with BNS to alleviate the domain gaps as follows:
\begin{align}
\setlength{\abovedisplayskip}{3pt}
\setlength{\belowdisplayskip}{3pt}
    \mathcal{L}_{bns} = & {|| \mu(F) - \bar{\mu}||}_2^2 + {|| \sigma^2(F) - \bar{\sigma}^2||}_2^2  \notag \\  
    &+ {|| \mu(F') - \bar{\mu}||}_2^2 + {|| \sigma^2(F') - \bar{\sigma}^2||}_2^2,    
 \end{align}
where $\bar{\mu}$ and $\bar{\sigma}^2$ are the mean and variance parameters of the last BN layer in the source model $S$.

% \begin{figure}
%     \centering
%     \includegraphics[width=\linewidth]{images/CDAM.pdf}
%     \caption{Illustration of cross dual attention module.}
%     \label{fig:cdam}
% \end{figure}

As shown in Fig.~\ref{fig:framework} \textcolor{red}{(a)}, after aligned with BNS, the ERP feature $f$ and the rebuilt feature $f'$ are first reshaped to be $f \in \mathbb{R}^{N \times C}$ and $f' \in \mathbb{R}^{N \times C}$, where $N$ is the number of pixels and $C$ is the channel number. Then we calculate the spatial-wise attention maps $M_{sp} \in \mathbb{R}^{N \times C}$ and ${M'}_{sp} \in \mathbb{R}^{N \times C}$ for $f$ and $f'$ by:
\begin{align}
\setlength{\abovedisplayskip}{3pt}
\setlength{\belowdisplayskip}{3pt}
    & \{M_{sp}\}_{ji} = \frac{exp({f'}_{[i:]} \cdot f_{[:j]}^T)}{\sum_i^{N} exp({f'}_{[i:]} \cdot f_{[:j]}^T)}, \qquad \notag \\ 
    & \{M_{sp}'\}_{ji} = \frac{exp(f_{[i:]} \cdot {f'}_{[:j]}^T)}{\sum_i^{N} exp(f_{[i:]} \cdot {f'}_{[:j]}^T)}, 
\end{align}
where $f^T$ is the transpose of $f$ and $\{M\}_{ij}$ measures the impact of the $i$-th position on the $j$-th position. 
Similarly, the channel-wise attention maps $M_{ch} \in \mathbb{R}^{C \times C}$ and ${M'}_{ch} \in \mathbb{R}^{C \times C}$ 
can be obtained through:
\begin{align}
    &\{M_{ch}\}_{ji} = \frac{exp({f'}_{[i:]}^T \cdot f_{[:j]})}{\sum_i^{C} exp({f'}_{[i:]} \cdot f_{[:j]}^T)}, \qquad \notag \\
    &\{M_{ch}'\}_{ji} = \frac{exp(f_{[i:]}^T \cdot {f'}_{[:j]})}{\sum_i^{C} exp(f_{[i:]}^T \cdot {f'}_{[:j]})}.
\end{align}
After obtaining the spatial and channel attention maps, the CDAM loss can be calculated with the Kullback-Liibler divergence (KL divergence) as follows:
\begin{equation}
    \mathcal{L}_{cda} = KL(M_{sp}, M_{sp}') + KL(M_{ch}, M_{ch}')
\end{equation}

\subsection{Optimization}
The training objective for learning the target model containing three losses is defined as:
\begin{equation}
    \mathcal{L} = \lambda \cdot \mathcal{L}_{ppa} + \gamma \cdot \mathcal{L}_{cda} + \mathcal{L}_{bns} + \mathcal{L}_{sup}
\end{equation}
where $\mathcal{L}_{ppa}$ is the MSE loss from PPAM, $\mathcal{L}_{cda}$ refers to the KL loss from CDAM, $\mathcal{L}_{sup}$ denotes the CE loss for the prediction pseudo label supervision loss, $\mathcal{L}_{bns}$ refers to the BNS guided feature loss, and $\lambda$ and $\gamma$ are the trade-off weights of the proposed loss terms. 
%And $\mathcal{L}_{sft}$ is for fine-tuning the source model.

\section{Experiments and Analysis}
As the first SFUDA method for panoramic image segmentation, there is no prior method for direct comparison. We thus empirically validate our method by comparing it with the existing UDA and panoramic segmentation methods on three widely used benchmarks.
% and show the experimental settings of our SFUDA for 
% panoramic semantic segmentation, and implementation details.
% We further provide the experimental results of our proposed SFUDA method compared with the existing UDA and panoramic segmentation methods on two widely applied benchmarks.

\subsection{Datasets and Implementation Details.}
Cityscapes~\cite{Cityscapes} is a real-world dataset collected for autonomous driving that contains street scenes. %captured from 50 different cities 
%with precise pixel-wise annotations of 19 semantic categories. 
%The official split contains 2975 images for training and 500 images for validation. 
%In this paper, we utilize the official training set as the source domain data for obtaining the accessible source model. 
DensePASS~\cite{densepass} is a panoramic dataset designed for capturing diverse street scenes.
%from all around the globe and contains 2500 panoramas. The panoramas are precisely annotated for the critical navigation-related categories as the test set. 
SynPASS~\cite{zhang2022behind} is a synthetic dataset consisting of 9080 synthetic panoramic images. 
%annotated with 22 categories. The official training, validation, and test sets contain 5700, 1690, and 1690 images, respectively.
%We use the overlapping 13 classes between SynPASS and DensePASS datasets for training and testing.
Stanford2D3D~\cite{armeni2017joint} is an indoor panoramic dataset which has 1413 panoramic images. %The images are annotated with the same 13 categories as its pinhole dataset.
Overall, the experiments are conducted on both real-world (Cityscapes-to-DensePASS, C-to-D, and Stanford2D3D-pinhole-to-Stanford2D3D-panoramic, SPin-to-SPan) and synthetic-to-real (SynPASS-to-DensePASS, S-to-D) scenarios. 
% \textit{More details, such as the implementation of tangent projection can be found in the supplementray material}.
%All the experiments in ablation study are conducted with SegFormer-B1 (Seg-B1) model.
\begin{table*}[]
\centering
\label{tab:stanford}
\setlength{\tabcolsep}{8pt}
\resizebox{0.99\linewidth}{!}{
\begin{tabular}{l|c|c|cccccccc|c}
\toprule
Method & SF & mIoU & Ceiling & Chair & Door & Floor & Sofa & Table & Wall & Window & $\Delta$ \\ \midrule
% PVT-S~\cite{zhang2022bending} & \XSolidBrush & 57.71 & 85.69 & 51.71 & 18.54 & 90.78 & 34.76 & 65.34 & 74.87 & 39.98 & - \\
PVT-S w/ MPA~\cite{zhang2022bending} & \XSolidBrush & 57.95 & 85.85 & 51.76 & 18.39 & 90.78 & 35.93 & 65.43 & 75.00 & 40.43 & - \\
Trans4PASS w/ MPA~\cite{zhang2022bending} & \XSolidBrush & 64.52 & 85.08 & 58.72 & 34.97 & 91.12 & 46.25 & 71.72 & 77.58 & 50.75 & - \\
Trans4PASS+~\cite{zhang2022behind} & \XSolidBrush & 63.73 & 90.63 & 62.30 & 24.79 & 92.62 & 35.73 & 73.16 & 78.74 & 51.78 & - \\
Trans4PASS+ w/ MPA~\cite{zhang2022behind} & \XSolidBrush & 67.16 & 90.04 & 64.04 & 42.89 & 91.74 & 38.34 & 71.45 & 81.24 & 57.54 & -
\\ \midrule
SFDA~\cite{liu2021source}  & \Checkmark & 54.76 & 79.44 & 33.20 & 52.09 & 67.36 & 22.54 & 53.64 & 69.38 & 60.46 & -\\
Ours w/ b1 & \Checkmark & 57.63 & 73.81 & 29.98 & 63.65 & 73.49 & 31.76 & 49.25 & 72.89 & 66.22 & \textcolor{red}{\textbf{+2.87}}\\ 
Ours w/ b2 & \Checkmark & 65.75 & 82.88 & 38.00 & \textbf{65.81} & 86.71 & 36.32 & 66.10 & 80.29 & \textbf{69.88} & \textcolor{red}{\textbf{+10.99}}\\
\bottomrule
\end{tabular}}
\caption{Experimental results on indoor Stanford2D3D~\cite{armeni2017joint}. The \textbf{bold} denotes the best performance among UDA and SFUDA methods.}
\end{table*}
\begin{table}[]
%% \vspace{-10pt}
\centering
\setlength{\tabcolsep}{4pt}
\resizebox{0.48\textwidth}{!}{
\begin{tabular}{ccccccccc}
\toprule
\multicolumn{5}{c}{Loss Function Combinations} & \multicolumn{2}{c}{C-to-D} & \multicolumn{2}{c}{S-to-D} \\ \midrule
$\mathcal{L}_{sup}$ & $\mathcal{L}_{ppa}$ & $\mathcal{L}_{sft}$ & $\mathcal{L}_{cda}$ & $\mathcal{L}_{bns}$ & mIoU & $\Delta$ & mIoU & $\Delta$ \\ \midrule
\Checkmark &  & &  &  & 38.65 & -  & 35.81 & - \\ 
\Checkmark & \Checkmark &  & &  & 45.42 & +6.77 & 38.37 & +2.56 \\ 
\Checkmark & \Checkmark & \Checkmark & &  & 46.23 & +7.58  & 38.49 & +2.68 \\ 
\Checkmark & & & \Checkmark &  & 44.24 & +5.59 & 38.38 & +2.57 \\ 
\Checkmark & & & \Checkmark & \Checkmark & 44.79 & +6.14 & 38.52 & +2.71  \\ 
\Checkmark & \Checkmark & \Checkmark & \Checkmark & \Checkmark & 48.78 & +10.13 & 41.78 & +5.97 \\ 
\bottomrule
\end{tabular}}
% \vspace{-8pt}
\caption{Ablation study of different module combinations.}
\vspace{-10pt}
\label{LossCombin}
\end{table}
\subsection{Experimental Results.}
\begin{figure}[t!]
    \centering
    \includegraphics[width=\linewidth]{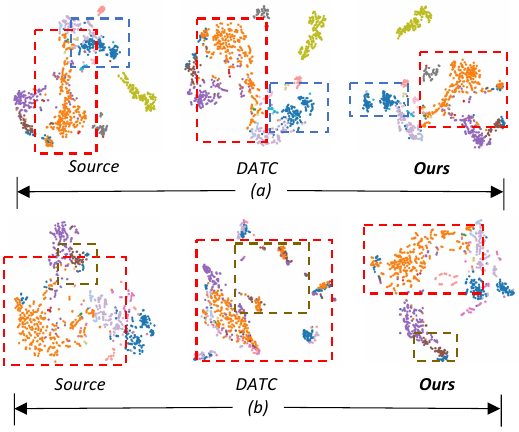}
    % \vspace{-20pt}
    \caption{TSNE visualization of (a) Cityscapes-to-DensePASS and (b) SynPASS-to-DensePASS.}
    \label{fig:tsne}
    % \vspace{-12pt}
\end{figure}

We first evaluate our proposed framework under the S-to-D scenario. The experimental results are shown in Tab.~\ref{Tab:syn2dense}. Our proposed method consistently outperforms source-free UDA methods ~\cite{liu2021source} and ~\cite{yang2022source} and even achieves panoramic semantic segmentation performance closer to that of the UDA method Trans4PASS ~\cite{zhang2022behind} which utilizes the source data in the adaptation procedure. Our proposed method brings significant performance gain of +3.57\% and +3.54\% with SegFormer-B1 backbone then SFDA ~\cite{liu2021source} and DATC~\cite{yang2022source}, respectively. We also provide the TSNE visualization in Fig.~\ref{fig:tsne} \textcolor{red}{(b)} and qualitative results in Fig.~\ref{fig:visualization}. Apparently, our method gains a significant improvement in distinguishing the pixels in panoramic images in both prediction and high-level feature space. 
As shown in Tab.~\ref{Tab:City2PASS}, we then evaluate our proposed framework under the C-to-D scenario. Our proposed method significantly outperforms source-free methods~\cite{liu2021source, yang2022source} and some panoramic semantic segmentation methods~\cite{p2pda, yang2020omnisupervised, PCS}. Specifically, our method achieves a significant performance gain over SFDA~\cite{liu2021source} and DTAC~\cite{yang2022source} by +6.08\% and +5.72\%, respectively. This demonstrates that our proposed method endowed by PPAM and CDAM is more suitable for panoramic semantic segmentation tasks. Furthermore, as shown in the qualitative results in Fig.~\ref{fig:visualization}, our method achieves better segmentation in driving-related categories, such as rider and car. 

We also provide TSNE visualizations~\cite{van2008visualizing} in Fig.~\ref{fig:tsne} (a), showing that our proposed method brings significant improvements in distinguishing pixels from different categories in high-level feature space.
Additionally, we evaluated our proposed method on the Stanford2D3D~\cite{armeni2017joint} dataset and compared it with the SFDA~\cite{liu2021source} and MPA~\cite{zhang2022behind} methods. As shown in the following table, our proposed method significantly outperforms the SFDA by +7.09\% mIoU and is on par with the MPA method using source data (61.85\% vs. 67.16\%). Notably, for some categories, such as door (57.90\% vs. 42.89\%) and window (68.06\% vs. 57.54\%), our method event outperforms the MPA~\cite{zhang2022behind}.
%% \vspace{-10pt}

\begin{table}[]
\centering
\setlength{\tabcolsep}{14pt}
\resizebox{\linewidth}{!}{
\begin{tabular}{cccc}
\toprule
Combinations & $\tau_g$+$\tau_p$  & $\tau_g$+$\tau_f$ & $\tau_g$+$\tau_p$+$\tau_f$  \\ \midrule
mIoU & 44.14 & 44.28 & \textbf{45.42}    \\ \bottomrule
\end{tabular}}
% \vspace{-8pt}
\caption{Ablation study of different prototype combinations.}
\vspace{-4pt}
\label{ProtoCombin}
\end{table}
\section{Ablation Study}
%% \vspace{-10pt}
\textbf{Different Loss Function Combinations.}
To assess the effectiveness of the proposed modules, we conduct ablation experiments on both real-world and synthetic-to-real scenarios with various loss combinations. All of the proposed modules and loss functions have a positive impact on improving segmentation performance. Notably, our PPAM yields a significant performance gain of +6.77\%. This indicates that PPAM alleviates the intricate semantics and distortion problem with the tangent, and our proposed FFP projection is valid. This is further supported by the qualitative results presented in Fig.~\ref{fig:visualization}. Additionally, our proposed CDAM achieves a performance gain of +5.59\% compared to the source baseline, which means that CDAM imitates the spatial and channel-wise distributions of ERP and FFP features and further addresses the style discrepancy problems.

\noindent \textbf{Ablation of Different Prototype Combinations.}
\label{Sec:ab}
To validate the effectiveness of all the prototypes in PPAM, we conduct experiments on C-to-D using SegFormer-B1 and only $\mathcal{L}_{sup}$ and $\mathcal{L}_{ppa}$. The results of the performance with different prototype combinations are presented in Tab.~\ref{ProtoCombin}. Both prototypes from TP and FFP have a positive effect on PPAM, with $\tau_p$ and $\tau_f$ resulting in mIoU improvements of +5.49\% and +5.63\%, respectively, compared to the source baseline. When both prototypes are combined together, there is a mIoU gain of +6.77\%, indicating that their combination is better for prototype-level adaptation.

% %% \vspace{-10pt}
\noindent\textbf{Dual Attention vs. Cross Dual Attention.}
The dual attention (DA) approach proposed in SFDA~\cite{liu2021source} aligns the spatial and channel characteristics of features between the fake source and target data. In contrast, our cross dual attention (CDA) approach aligns the distribution between different projections of the same spherical data, specifically ERP and FFP, resulting in more robust and stable knowledge transfer. Moreover, in our SFDA, we obtain spatial and channel characteristics across features, whereas DA operates within features. We also evaluate DA on the C-to-D scenario, and our CDA achieves 44.24\% mIoU, while DA only reaches 41.53\% mIoU. This indicates the proposed CDA is better for SFUDA in panoramic semantic segmentation.

\noindent \textbf{Field-of-view of FFP.}
Most existing approaches for panoramic semantic segmentation, such as those proposed in \cite{zhang2022bending, zhang2022behind, zheng2023both}, primarily focus on alleviating distortion by introducing distortion-aware components and distinct projection strategies. However, as discussed in Sec.~\ref{Sec.3.2}, $360^\circ$ images contain more intricate semantic information and object correspondence than the pinhole images, resulting in an obvious semantic mismatch between domains. Therefore, we propose the Fixed FoV Pooling (FFP) strategy to address the semantic mismatch. Experimental results show that the fixed FoV is the most influential factor in FFP, with an FoV of $90^\circ$ achieving the best segmentation performance, as shown in Tab.~\ref{FFPFoV}, with a mIoU of 44.28\%.

\begin{table}[]
\setlength{\tabcolsep}{6pt}
\resizebox{0.49\textwidth}{!}{
\begin{tabular}{cccccccc}
\toprule
FoV & w/o & $60^\circ$ & $72^\circ$ & $90^\circ$ & $120^\circ$ & $180^\circ$ & $360^\circ$ \\ \midrule
mIoU & 38.65 & 44.03 & 44.16 & \textbf{44.28} & 44.02 & 41.65 & 40.31 \\ 
$\Delta$ & - & +5.38 & +5.51 & \textbf{+5.63} & +5.37 & +3.00 & +1.66 \\
\bottomrule
\end{tabular}}
% \vspace{-8pt}
\caption{Ablation study of the FoV of our proposed FFP.}
% \vspace{-8pt}
%% \vspace{-8pt}
\label{FFPFoV}
\end{table}

\begin{table}[]
\centering
\resizebox{0.49\textwidth}{!}{
\setlength{\tabcolsep}{8pt}
\begin{tabular}{ccccccc}
\toprule
$\gamma$ & 0 & 0.01 & 0.02 & 0.05 & 0.1 & 0.2 \\ \midrule
        mIoU & 38.65 & 42.05 & 43.24 & 43.28 & \textbf{44.24} & 43.07 \\ 
        $\Delta$ & - & +3.40 & +4.59 & +4.63 & \textbf{+5.59} & +4.42 \\
\bottomrule
\end{tabular}}
\resizebox{0.49\textwidth}{!}{
\setlength{\tabcolsep}{4pt}
\begin{tabular}{ccccccccc}
\toprule
$\lambda$ & 0 & 50 & 60 & 80& 100 & 120 & 150 & 200\\ \midrule
        mIoU & 38.65 & 43.13 & 43.22 & 45.36 & \textbf{45.42} & 45.34 & 45.33 & 45.12 \\ 
        $\Delta$ & - & +4.48 & +4.57 & +6.71 & \textbf{+6.77} & +6.69 & +6.68 & +6.47 \\
\bottomrule
\end{tabular}}
% \vspace{-8pt}
\caption{Ablation study of $\gamma$ and $\lambda$.}
\vspace{-6pt}
\label{Tab:gammaab}
\end{table}

% \begin{table}[]
% \caption{Ablation study of $\lambda$.}
% \setlength{\tabcolsep}{4pt}
% \resizebox{0.49\textwidth}{!}{
% \begin{tabular}{ccccccccc}
% \toprule
% $\lambda$ & 0 & 50 & 60 & 80& 100 & 120 & 150 & 200\\ \midrule
%         mIoU & 38.65 & 43.13 & 43.22 & 45.36 & 45.42 & 45.34 & 45.33 & 45.12 \\ 
%         $\Delta$ & - & +4.48 & +4.57 & +6.71 & \textbf{+6.77} & +6.69 & +6.68 & +6.47 \\
% \bottomrule
% \end{tabular}}
% \label{Tab:lambdaab}
% \end{table}

% \noindent \textbf{Update Strategy of the Panoramic Global Prototypes.}
% As outlined in Sec.~\ref{Sec.3.2}, our panoramic prototypes for adaptation are updated iteratively throughout the training process. To demonstrate the effectiveness of our iterative strategy in Eq.~(\ref{Eq:iter}), we substitute it with a non-iterative approach. We conduct experiments on the C-to-D scenario with the SegFormer-B1 model, using $\mathcal{L}{sup}$ and $\mathcal{L}{ppa}$ with both the non-iterative and iterative approaches. The resulting mIoU values are \textbf{43.08\%} and \textbf{45.42\%}, respectively. Our results show that the iterative update makes the panoramic prototypes more robust and holistic.

\noindent \textbf{Ablation of Hyper-parameters.}
%We now discuss the 
We now show the influence of hyperparameters $\gamma$ and $\lambda$, which are the weights for the KL loss in CDAM and the MSE loss in PPAM, respectively. The experimental results are provided in Tab.~\ref{Tab:gammaab}. % for $\gamma$ and $\lambda$. 

%% \vspace{-10pt}
% \section{Discussion}
%% \vspace{-10pt}

\noindent \textbf{Fine-tuning the Source Model.}
As the pre-trained model in the source (pinhole) domain is not an ideal model for the target (panoramic) image domain, we propose to fine-tune the source model with the loss function $\mathcal{L}_{sft}$, as described in Sec.~\ref{Sec.3.2}. Tab.~\ref{LossCombin} demonstrates the effectiveness of the proposed $\mathcal{L}_{sft}$. When combined with the prototypical adaptation loss $\mathcal{L}_{ppa}$, adding $\mathcal{L}_{sft}$ results in a 6.77\% mIoU gain compared with the source baseline of 38.65\%. 
We present the performance metrics derived solely from the loss $\mathcal{L}_{sft}$ of PPAM: C-2-D registers at 44.94\% while S-2-D records 36.74\%. These results underscore the efficacy of $\mathcal{L}_{sft}$ integrated within our PPAM module. Concerning transfer-ability, our $\mathcal{L}_{sft}$ exhibits compatibility with various projection methods, \eg, cube map. At its core, our fine-tuning loss seeks to align all projection images originating from the same panoramic source, irrespective of the employed projection technique. This intrinsic adaptability facilitates the application of $\mathcal{L}_{sft}$ across diverse projections.
\textit{More results refer to the supplementary material.}
\section{Conclusion}
%% \vspace{-10pt}

In this paper, we investigated a new problem of achieving SFUDA for panoramic semantic segmentation. To this end, we proposed an end-to-end SFUDA framework to address the domain shifts, including semantic mismatch, distortion, and style discrepancies, between pinhole and panoramic domains.Experiments on both real-world and synthetic benchmarks show that our proposed framework outperforms prior approaches and is on par with the methods using source data. 

\noindent \textbf{Limitation and future work.}
One limitation of our proposed framework is the computational cost brought by the tangent projection during training, and there is still room for improvements in segmentation performance. However, components in our approach such as panoramic prototypes and fixed FoV projection have significant implications for the $360^\circ$ vision, especially for the panoramic semantic segmentation. 
In the future, we plan to utilize the large language models (LLMs) and Multi-modal large language models (MLLMs) to alleviate the domain gaps, such as the semantic mismatches between pinhole and panoramic images.
%In the future, we plan to extend our approach based on the recent advance in large vision models and introduce text supervision into panoramic image domain with the large language models.\\
% \noindent \textbf{Broader Impact}
% This paper serves as the first attempt to transfer knowledge from the pinhole image domain model to the panoramic images without accessing the pinhole images. The proposed SFUDA framework can be potentially used to adapt large visual models to the panoramic image domain and accelerate the applications of $360^\circ$ data.

\noindent \textbf{Acknowledgement} This paper is supported by the National Natural Science Foundation of China (NSF) under Grant No. NSFC22FYT45 and the Guangzhou City, University and Enterprise Joint Fund under Grant No.SL2022A03J01278.

\clearpage
{
    \small
    \bibliographystyle{ieeenat_fullname}
    \bibliography{main}
}

% WARNING: do not forget to delete the supplementary pages from your submission 
% \input{sec/X_suppl}

\end{document}